\PassOptionsToPackage{unicode}{hyperref}
\PassOptionsToPackage{hyphens}{url}
\PassOptionsToPackage{dvipsnames,svgnames,x11names}{xcolor}
\documentclass[
]{article}
\usepackage{amsmath,amssymb}
\usepackage{lmodern}
\usepackage{iftex}
\ifPDFTeX
  \usepackage[T1]{fontenc}
  \usepackage[utf8]{inputenc}
  \usepackage{textcomp} % provide euro and other symbols
\else % if luatex or xetex
  \usepackage{unicode-math}
  \defaultfontfeatures{Scale=MatchLowercase}
  \defaultfontfeatures[\rmfamily]{Ligatures=TeX,Scale=1}
\fi
\IfFileExists{upquote.sty}{\usepackage{upquote}}{}
\IfFileExists{microtype.sty}{% use microtype if available
  \usepackage[]{microtype}
  \UseMicrotypeSet[protrusion]{basicmath} % disable protrusion for tt fonts
}{}
\makeatletter
\@ifundefined{KOMAClassName}{% if non-KOMA class
  \IfFileExists{parskip.sty}{%
    \usepackage{parskip}
  }{% else
    \setlength{\parindent}{0pt}
    \setlength{\parskip}{6pt plus 2pt minus 1pt}}
}{% if KOMA class
  \KOMAoptions{parskip=half}}
\makeatother
\usepackage{xcolor}
\usepackage{color}
\usepackage{fancyvrb}

\DefineVerbatimEnvironment{Highlighting}{Verbatim}{commandchars=\\\{\}}
\newenvironment{Shaded}{}{}

\newcommand{\BuiltInTok}[1]{\textcolor[rgb]{0.00,0.50,0.00}{#1}}

\newcommand{\ConstantTok}[1]{\textcolor[rgb]{0.53,0.00,0.00}{#1}}
\newcommand{\ControlFlowTok}[1]{\textcolor[rgb]{0.00,0.44,0.13}{\textbf{#1}}}
\newcommand{\DataTypeTok}[1]{\textcolor[rgb]{0.56,0.13,0.00}{#1}}
\newcommand{\DecValTok}[1]{\textcolor[rgb]{0.25,0.63,0.44}{#1}}

\newcommand{\KeywordTok}[1]{\textcolor[rgb]{0.00,0.44,0.13}{\textbf{#1}}}
\newcommand{\NormalTok}[1]{#1}
\newcommand{\OperatorTok}[1]{\textcolor[rgb]{0.40,0.40,0.40}{#1}}

\newcommand{\PreprocessorTok}[1]{\textcolor[rgb]{0.74,0.48,0.00}{#1}}

\providecommand{\tightlist}{%
  \setlength{\itemsep}{0pt}\setlength{\parskip}{0pt}}
\NewDocumentCommand\citeproctext{}{}
\NewDocumentCommand\citeproc{mm}{%
  \begingroup\def\citeproctext{#2}\cite{#1}\endgroup}
\makeatletter
 \let\@cite@ofmt\@firstofone
 \def\@biblabel#1{}
 \def\@cite#1#2{{#1\if@tempswa , #2\fi}}
\makeatother
\newlength{\cslhangindent}
\newlength{\csllabelwidth}
\newenvironment{CSLReferences}[2] % #1 hanging-indent, #2 entry-spacing
 {\begin{list}{}{%
  \setlength{\itemindent}{0pt}
  \setlength{\leftmargin}{0pt}
  \setlength{\parsep}{0pt}
  \ifodd #1
   \setlength{\leftmargin}{\cslhangindent}
   \setlength{\itemindent}{-1\cslhangindent}
  \fi
  \setlength{\itemsep}{#2\baselineskip}}}
 {\end{list}}
\usepackage{calc}

\ifLuaTeX
\usepackage[bidi=basic]{babel}
\else
\usepackage[bidi=default]{babel}
\fi
\babelprovide[main,import]{american}
\def\languageshorthands#1{}
\ifLuaTeX
  \usepackage{selnolig}  % disable illegal ligatures
\fi
\IfFileExists{bookmark.sty}{\usepackage{bookmark}}{\usepackage{hyperref}}
\IfFileExists{xurl.sty}{\usepackage{xurl}}{} % add URL line breaks if available
\hypersetup{
  pdftitle={Basin: Efficient and Extensible Numerical Optimization in
Rust},
  pdfauthor={Johan Larsson},
  pdflang={en-US},
  colorlinks=true,
  linkcolor={Maroon},
  filecolor={Maroon},
  citecolor={Blue},
  urlcolor={Blue},
  pdfcreator={LaTeX via pandoc}}

\title{Basin: Efficient and Extensible Numerical Optimization in Rust}

\definecolor{c53baa1}{RGB}{83,186,161}
\definecolor{c202826}{RGB}{32,40,38}
\def \rorglobalscale {0.1}
\newcommand{\rorlogo}{%
\begin{tikzpicture}[y=1cm, x=1cm, yscale=\rorglobalscale,xscale=\rorglobalscale, every node/.append style={scale=\rorglobalscale}, inner sep=0pt, outer sep=0pt]
  \begin{scope}[even odd rule,line join=round,miter limit=2.0,shift={(-0.025, 0.0216)}]
    \path[fill=c53baa1,nonzero rule,line join=round,miter limit=2.0] (1.8164, 3.012) -- (1.4954, 2.5204) -- (1.1742, 3.012) -- (1.8164, 3.012) -- cycle;
    \path[fill=c53baa1,nonzero rule,line join=round,miter limit=2.0] (3.1594, 3.012) -- (2.8385, 2.5204) -- (2.5172, 3.012) -- (3.1594, 3.012) -- cycle;
    \path[fill=c53baa1,nonzero rule,line join=round,miter limit=2.0] (1.1742, 0.0669) -- (1.4954, 0.5588) -- (1.8164, 0.0669) -- (1.1742, 0.0669) -- cycle;
    \path[fill=c53baa1,nonzero rule,line join=round,miter limit=2.0] (2.5172, 0.0669) -- (2.8385, 0.5588) -- (3.1594, 0.0669) -- (2.5172, 0.0669) -- cycle;
    \path[fill=c202826,nonzero rule,line join=round,miter limit=2.0] (3.8505, 1.4364).. controls (3.9643, 1.4576) and (4.0508, 1.5081) .. (4.1098, 1.5878).. controls (4.169, 1.6674) and (4.1984, 1.7642) .. (4.1984, 1.8777).. controls (4.1984, 1.9719) and (4.182, 2.0503) .. (4.1495, 2.1132).. controls (4.1169, 2.1762) and (4.0727, 2.2262) .. (4.0174, 2.2635).. controls (3.9621, 2.3006) and (3.8976, 2.3273) .. (3.824, 2.3432).. controls (3.7505, 2.359) and (3.6727, 2.367) .. (3.5909, 2.367) -- (2.9676, 2.367) -- (2.9676, 1.8688).. controls (2.9625, 1.8833) and (2.9572, 1.8976) .. (2.9514, 1.9119).. controls (2.9083, 2.0164) and (2.848, 2.1056) .. (2.7705, 2.1791).. controls (2.6929, 2.2527) and (2.6014, 2.3093) .. (2.495, 2.3487).. controls (2.3889, 2.3881) and (2.2728, 2.408) .. (2.1468, 2.408).. controls (2.0209, 2.408) and (1.905, 2.3881) .. (1.7986, 2.3487).. controls (1.6925, 2.3093) and (1.6007, 2.2527) .. (1.5232, 2.1791).. controls (1.4539, 2.1132) and (1.3983, 2.0346) .. (1.3565, 1.9436).. controls (1.3504, 2.009) and (1.3351, 2.0656) .. (1.3105, 2.1132).. controls (1.2779, 2.1762) and (1.2338, 2.2262) .. (1.1785, 2.2635).. controls (1.1232, 2.3006) and (1.0586, 2.3273) .. (0.985, 2.3432).. controls (0.9115, 2.359) and (0.8337, 2.367) .. (0.7519, 2.367) -- (0.1289, 2.367) -- (0.1289, 0.7562) -- (0.4837, 0.7562) -- (0.4837, 1.4002) -- (0.6588, 1.4002) -- (0.9956, 0.7562) -- (1.4211, 0.7562) -- (1.0118, 1.4364).. controls (1.1255, 1.4576) and (1.2121, 1.5081) .. (1.2711, 1.5878).. controls (1.2737, 1.5915) and (1.2761, 1.5954) .. (1.2787, 1.5991).. controls (1.2782, 1.5867) and (1.2779, 1.5743) .. (1.2779, 1.5616).. controls (1.2779, 1.4327) and (1.2996, 1.3158) .. (1.3428, 1.2113).. controls (1.3859, 1.1068) and (1.4462, 1.0176) .. (1.5237, 0.944).. controls (1.601, 0.8705) and (1.6928, 0.8139) .. (1.7992, 0.7744).. controls (1.9053, 0.735) and (2.0214, 0.7152) .. (2.1474, 0.7152).. controls (2.2733, 0.7152) and (2.3892, 0.735) .. (2.4956, 0.7744).. controls (2.6016, 0.8139) and (2.6935, 0.8705) .. (2.771, 0.944).. controls (2.8482, 1.0176) and (2.9086, 1.1068) .. (2.952, 1.2113).. controls (2.9578, 1.2253) and (2.9631, 1.2398) .. (2.9681, 1.2544) -- (2.9681, 0.7562) -- (3.3229, 0.7562) -- (3.3229, 1.4002) -- (3.4981, 1.4002) -- (3.8349, 0.7562) -- (4.2603, 0.7562) -- (3.8505, 1.4364) -- cycle(0.9628, 1.7777).. controls (0.9438, 1.7534) and (0.92, 1.7357) .. (0.8911, 1.7243).. controls (0.8623, 1.7129) and (0.83, 1.706) .. (0.7945, 1.7039).. controls (0.7588, 1.7015) and (0.7252, 1.7005) .. (0.6932, 1.7005) -- (0.4839, 1.7005) -- (0.4839, 2.0667) -- (0.716, 2.0667).. controls (0.7477, 2.0667) and (0.7805, 2.0643) .. (0.8139, 2.0598).. controls (0.8472, 2.0553) and (0.8768, 2.0466) .. (0.9025, 2.0336).. controls (0.9282, 2.0206) and (0.9496, 2.0021) .. (0.9663, 1.9778).. controls (0.9829, 1.9534) and (0.9914, 1.9209) .. (0.9914, 1.8799).. controls (0.9914, 1.8362) and (0.9819, 1.8021) .. (0.9628, 1.7777) -- cycle(2.6125, 1.3533).. controls (2.5889, 1.2904) and (2.5553, 1.2359) .. (2.5112, 1.1896).. controls (2.4672, 1.1433) and (2.4146, 1.1073) .. (2.3529, 1.0814).. controls (2.2916, 1.0554) and (2.2228, 1.0427) .. (2.1471, 1.0427).. controls (2.0712, 1.0427) and (2.0026, 1.0557) .. (1.9412, 1.0814).. controls (1.8799, 1.107) and (1.8272, 1.1433) .. (1.783, 1.1896).. controls (1.7391, 1.2359) and (1.7052, 1.2904) .. (1.6817, 1.3533).. controls (1.6581, 1.4163) and (1.6465, 1.4856) .. (1.6465, 1.5616).. controls (1.6465, 1.6359) and (1.6581, 1.705) .. (1.6817, 1.7687).. controls (1.7052, 1.8325) and (1.7388, 1.8873) .. (1.783, 1.9336).. controls (1.8269, 1.9799) and (1.8796, 2.0159) .. (1.9412, 2.0418).. controls (2.0026, 2.0675) and (2.0712, 2.0804) .. (2.1471, 2.0804).. controls (2.223, 2.0804) and (2.2916, 2.0675) .. (2.3529, 2.0418).. controls (2.4143, 2.0161) and (2.467, 1.9799) .. (2.5112, 1.9336).. controls (2.5551, 1.8873) and (2.5889, 1.8322) .. (2.6125, 1.7687).. controls (2.636, 1.705) and (2.6477, 1.6359) .. (2.6477, 1.5616).. controls (2.6477, 1.4856) and (2.636, 1.4163) .. (2.6125, 1.3533) -- cycle(3.8015, 1.7777).. controls (3.7825, 1.7534) and (3.7587, 1.7357) .. (3.7298, 1.7243).. controls (3.701, 1.7129) and (3.6687, 1.706) .. (3.6333, 1.7039).. controls (3.5975, 1.7015) and (3.5639, 1.7005) .. (3.5319, 1.7005) -- (3.3226, 1.7005) -- (3.3226, 2.0667) -- (3.5547, 2.0667).. controls (3.5864, 2.0667) and (3.6192, 2.0643) .. (3.6526, 2.0598).. controls (3.6859, 2.0553) and (3.7155, 2.0466) .. (3.7412, 2.0336).. controls (3.7669, 2.0206) and (3.7883, 2.0021) .. (3.805, 1.9778).. controls (3.8216, 1.9534) and (3.8301, 1.9209) .. (3.8301, 1.8799).. controls (3.8301, 1.8362) and (3.8206, 1.8021) .. (3.8015, 1.7777) -- cycle;
  \end{scope}
\end{tikzpicture}
}

\usepackage[affil-it]{authblk}
\usepackage{orcidlink}
\author[1%
  ]{Johan Larsson%
    \,\orcidlink{0000-0002-4029-5945}\,%
    }

\affil[1]{Department of Mathematical Sciences, University of Copenhagen,
Denmark%
    \,\protect\href{https://ror.org/035b05819}{\protect\rorlogo}\,%
  }
\date{11 July 2026}

\begin{document}
\maketitle

\section{Summary}\label{summary}

Basin is a numerical optimization library for the
\href{https://www.rust-lang.org}{Rust} programming
language~(\citeproc{ref-matsakis2014}{Matsakis \& Klock, 2014}).
Numerical optimization is the task of finding the inputs that minimize a
function, and it is a fundamental element across the sciences: fitting a
model to data, calibrating a simulation, training a machine learning
model, or choosing engineering parameters that minimize cost. Basin
gives users a single, consistent way to both state and solve such
problems, with a broad catalog of solvers and first-class support for
constraints.

To use Basin, a user implements one or more small traits describing
their objective---at minimum a \texttt{CostFunction} that returns a
value for a given input, and optionally its derivatives
(\texttt{Gradient}, \texttt{Jacobian}, or \texttt{Hessian}). The user
then hands the problem, a solver, and a starting point to an
\texttt{Executor}, which drives the optimization loop, handles stopping
criteria, and returns the result. Basin works out of the box on plain
Rust vectors and, optionally, with faster linear-algebra backends
available behind feature flags. The default build compiles to
WebAssembly, which means that Basin can be used in a browser without a
native toolchain or BLAS/LAPACK support. Documentation is published at
\href{https://basin.rs}{basin.rs}, which includes a user guide,
interactive visualizer, and a benchmark suite comparing Basin to other
optimization libraries.

\section{Statement of Need}\label{statement-of-need}

Rust is increasingly used for scientific and numerical computing because
it combines performance with memory safety and a strong package
ecosystem. Optimization, however, is fragmented across the ecosystem:
most crates specialize in a single family of methods and no widely used
Rust crate couples a broad solver catalog with first-class constraints
and a browser-ready default build. Basin was written to close that gap,
and it targets four concrete needs.

First, Basin includes a broad catalog of solvers. Real problems rarely
announce in advance what optimization algorithm they need, so Basin
includes a large set of solvers behind a single, consistent API. The
catalog includes

\begin{itemize}
\tightlist
\item
  first-order and quasi-Newton methods (gradient descent, SGD, BFGS,
  L-BFGS, L-BFGS-B, and a Newton trust-region
  method)~(\citeproc{ref-byrd1995}{Byrd et al., 1995};
  \citeproc{ref-nocedal2006}{Nocedal \& Wright, 2006};
  \citeproc{ref-zhu1997}{Zhu et al., 1997});
\item
  derivative-free methods
  (Nelder--Mead~(\citeproc{ref-nelder1965}{Nelder \& Mead, 1965}),
  one-dimensional Brent~(\citeproc{ref-brent2013}{Brent, 2013}) and
  golden-section searches, NEWUOA~(\citeproc{ref-powell2006}{Powell,
  2006}), BOBYQA~(\citeproc{ref-powell2009}{Powell, 2009}),
  LINCOA~(\citeproc{ref-powell2015}{Powell, 2015}),
  COBYLA~(\citeproc{ref-powell1994}{Powell, 1994}), and mesh adaptive
  direct search~(\citeproc{ref-audet2006}{Audet \& Dennis, 2006}));
\item
  nonlinear least squares (Gauss--Newton and
  Levenberg--Marquardt~(\citeproc{ref-nielsen1999}{Nielsen, 1999}));
\item
  global and stochastic methods (random
  search~(\citeproc{ref-brooks1958}{Brooks, 1958}),
  CMA-ES~(\citeproc{ref-hansen2016}{Hansen, 2016}), differential
  evolution~(\citeproc{ref-storn1997}{Storn \& Price, 1997}), a
  steady-state genetic algorithm~(\citeproc{ref-molina2010}{Molina et
  al., 2010}), and basin-hopping~(\citeproc{ref-wales1997}{Wales \&
  Doye, 1997})); and
\item
  memetic combinations (MA-LS-Chain~(\citeproc{ref-molina2010}{Molina et
  al., 2010}), plus CMA-ES and differential evolution injection
  wrappers).
\end{itemize}

Switching methods is simple, sometimes requiring changing only a single
line of code.

Second, the design enforces correctness at compile time. Solvers,
termination criteria, and observers in Basin bind on the minimum state
shape they require, which means that a method that exposes no gradient
cannot be paired with a gradient-based stopping rule---mismatches yield
compilation errors rather than runtime failures.

Third, Basin is designed to be portable. The default build targets
WebAssembly with neither BLAS/LAPACK nor concurrency dependencies, which
means that Basin can be run in a browser without a native toolchain. It
also supports a low minimum supported Rust version in order to
facilitate its use in R packages and other scientific programming
languages that can be extended through Rust.

Fourth, support for constraints is first-class. Constraints are declared
on the problem, not passed to the solver call, and a solver that
requires constraints will not accept an unconstrained problem---again a
compile error rather than a runtime one. Basin supports box bounds,
linear equality and inequality constraints, and nonlinear inequality
constraints, together with opt-in adapters (log-barrier and augmented
Lagrangian) that recast a constrained problem as an unconstrained one so
that any unconstrained solver can be applied to it.

The target audience is researchers, engineers, and students who need
reliable optimization in Rust or any of the scientific programming
languages that can be easily extended through Rust, such as R, Julia,
and Python.

\section{State of the Field}\label{state-of-the-field}

The closest analog to Basin is
argmin~(\citeproc{ref-kroboth2025}{Kroboth, 2018/2025}): a numerical
optimization framework from which Basin takes considerable inspiration,
including the \texttt{Executor} driver loop, the
\texttt{Solver}/\texttt{Problem} trait split, and per-solver
\texttt{State}. But Basin diverges elsewhere, bringing

\begin{itemize}
\tightlist
\item
  first-class, problem-side constraints rather than solver
  configuration,
\item
  a richer linear-algebra tier implemented in pure Rust, and
\item
  generic termination criteria shared between solvers.
\end{itemize}

gomez~(\citeproc{ref-nevyhosteny2025}{Nevyhoštěný, 2021/2025}) is
another Rust crate with similar scope, implementing a small set of
derivative-free methods and nonlinear least-squares solvers, and
supporting constraints. Compared to Basin, it has a smaller solver
catalog, only supports box constraints, and does not have a generic
backend tier for linear algebra.

Finally, the nlopt crate provides a Rust interface to the NLopt C
library~(\citeproc{ref-johnson2026}{Johnson \& Schueller, 2013/2026}).
Although NLopt has a broad catalog of solvers, it requires a C toolchain
to build and is not WebAssembly-compatible. It also lacks Basin's
generic termination criteria and first-class constraints support.

In summary, Basin's contribution is to bring a broad catalog natively to
Rust and WebAssembly, without linking a C or Fortran toolchain in its
default configuration.

\section{Example}\label{example}

In the following example, we implement the Rosenbrock function and its
gradient, then minimize it with gradient descent. The \texttt{Executor}
driver loop handles the iteration, stopping criteria, and error
handling.

\begin{Shaded}
\begin{Highlighting}[]
\KeywordTok{use} \PreprocessorTok{basin::}\OperatorTok{\{}
\NormalTok{    BasicState}\OperatorTok{,}\NormalTok{ CostFunction}\OperatorTok{,}\NormalTok{ Executor}\OperatorTok{,}\NormalTok{ Gradient}\OperatorTok{,}\NormalTok{ GradientDescent}\OperatorTok{,}
\NormalTok{    GradientTolerance}\OperatorTok{,}
\OperatorTok{\};}
\KeywordTok{use} \PreprocessorTok{std::convert::}\NormalTok{Infallible}\OperatorTok{;}

\KeywordTok{struct}\NormalTok{ Rosenbrock}\OperatorTok{;}

\KeywordTok{impl}\NormalTok{ CostFunction }\ControlFlowTok{for}\NormalTok{ Rosenbrock }\OperatorTok{\{}
    \KeywordTok{type}\NormalTok{ Param }\OperatorTok{=} \DataTypeTok{Vec}\OperatorTok{\textless{}}\DataTypeTok{f64}\OperatorTok{\textgreater{};}
    \KeywordTok{type}\NormalTok{ Output }\OperatorTok{=} \DataTypeTok{f64}\OperatorTok{;}
    \KeywordTok{type}\NormalTok{ Error }\OperatorTok{=}\NormalTok{ Infallible}\OperatorTok{;}

    \KeywordTok{fn}\NormalTok{ cost(}\OperatorTok{\&}\KeywordTok{self}\OperatorTok{,}\NormalTok{ x}\OperatorTok{:} \OperatorTok{\&}\DataTypeTok{Vec}\OperatorTok{\textless{}}\DataTypeTok{f64}\OperatorTok{\textgreater{}}\NormalTok{) }\OperatorTok{{-}\textgreater{}} \DataTypeTok{Result}\OperatorTok{\textless{}}\DataTypeTok{f64}\OperatorTok{,} \DataTypeTok{Self}\PreprocessorTok{::}\BuiltInTok{Error}\OperatorTok{\textgreater{}} \OperatorTok{\{}
        \ConstantTok{Ok}\NormalTok{((}\DecValTok{1.0} \OperatorTok{{-}}\NormalTok{ x[}\DecValTok{0}\NormalTok{])}\OperatorTok{.}\NormalTok{powi(}\DecValTok{2}\NormalTok{) }\OperatorTok{+} \DecValTok{100.0} \OperatorTok{*}\NormalTok{ (x[}\DecValTok{1}\NormalTok{] }\OperatorTok{{-}}\NormalTok{ x[}\DecValTok{0}\NormalTok{]}\OperatorTok{.}\NormalTok{powi(}\DecValTok{2}\NormalTok{))}\OperatorTok{.}\NormalTok{powi(}\DecValTok{2}\NormalTok{))}
    \OperatorTok{\}}
\OperatorTok{\}}

\KeywordTok{impl}\NormalTok{ Gradient }\ControlFlowTok{for}\NormalTok{ Rosenbrock }\OperatorTok{\{}
    \KeywordTok{type}\NormalTok{ Gradient }\OperatorTok{=} \DataTypeTok{Vec}\OperatorTok{\textless{}}\DataTypeTok{f64}\OperatorTok{\textgreater{};}

    \KeywordTok{fn}\NormalTok{ gradient(}\OperatorTok{\&}\KeywordTok{self}\OperatorTok{,}\NormalTok{ x}\OperatorTok{:} \OperatorTok{\&}\DataTypeTok{Vec}\OperatorTok{\textless{}}\DataTypeTok{f64}\OperatorTok{\textgreater{}}\NormalTok{) }\OperatorTok{{-}\textgreater{}} \DataTypeTok{Result}\OperatorTok{\textless{}}\DataTypeTok{Vec}\OperatorTok{\textless{}}\DataTypeTok{f64}\OperatorTok{\textgreater{},} \DataTypeTok{Self}\PreprocessorTok{::}\BuiltInTok{Error}\OperatorTok{\textgreater{}} \OperatorTok{\{}
        \ConstantTok{Ok}\NormalTok{(}\PreprocessorTok{vec!}\NormalTok{[}
            \OperatorTok{{-}}\DecValTok{2.0} \OperatorTok{*}\NormalTok{ (}\DecValTok{1.0} \OperatorTok{{-}}\NormalTok{ x[}\DecValTok{0}\NormalTok{]) }\OperatorTok{{-}} \DecValTok{400.0} \OperatorTok{*}\NormalTok{ x[}\DecValTok{0}\NormalTok{] }\OperatorTok{*}\NormalTok{ (x[}\DecValTok{1}\NormalTok{] }\OperatorTok{{-}}\NormalTok{ x[}\DecValTok{0}\NormalTok{]}\OperatorTok{.}\NormalTok{powi(}\DecValTok{2}\NormalTok{))}\OperatorTok{,}
            \DecValTok{200.0} \OperatorTok{*}\NormalTok{ (x[}\DecValTok{1}\NormalTok{] }\OperatorTok{{-}}\NormalTok{ x[}\DecValTok{0}\NormalTok{]}\OperatorTok{.}\NormalTok{powi(}\DecValTok{2}\NormalTok{))}\OperatorTok{,}
\NormalTok{        ])}
    \OperatorTok{\}}
\OperatorTok{\}}

\KeywordTok{fn}\NormalTok{ main() }\OperatorTok{\{}
    \KeywordTok{let}\NormalTok{ result }\OperatorTok{=} \PreprocessorTok{Executor::}\NormalTok{new(}
\NormalTok{        Rosenbrock}\OperatorTok{,}
        \PreprocessorTok{GradientDescent::}\NormalTok{new(}\DecValTok{1e{-}3}\NormalTok{)}\OperatorTok{,}
        \PreprocessorTok{BasicState::}\NormalTok{new(}\PreprocessorTok{vec!}\NormalTok{[}\OperatorTok{{-}}\DecValTok{1.2}\OperatorTok{,} \DecValTok{1.0}\NormalTok{])}\OperatorTok{,}
\NormalTok{    )}
    \OperatorTok{.}\NormalTok{max\_iter(}\DecValTok{50\_000}\NormalTok{)}
    \OperatorTok{.}\NormalTok{terminate\_on(GradientTolerance(}\DecValTok{1e{-}6}\NormalTok{))}
    \OperatorTok{.}\NormalTok{run()}
    \OperatorTok{.}\NormalTok{unwrap()}\OperatorTok{;}
\OperatorTok{\}}
\end{Highlighting}
\end{Shaded}

\section{Software Design}\label{software-design}

Basin is organized as a generic core with a broad catalog of solvers
layered on top of it. The design is built on a set of principles that we
think make it easy to extend and maintain.

\subsection{Tiered Backends}\label{tiered-backends}

Parameters and linear algebra are generic over the backend. A universal
\emph{vector tier} (operations such as scaled addition, dot products,
and norms that every backend implements well) keeps first-order and
derivative-free solvers backend-generic across
\texttt{Vec\textless{}f64\textgreater{}},
nalgebra~(\citeproc{ref-crozet2026}{Crozet, 2013/2026}),
ndarray~(\citeproc{ref-sverdrup2026}{Sverdrup \& Turner, 2014/2026}),
and faer~(\citeproc{ref-sarrazin2026}{Sarrazin, 2022/2026}). Each
backend is activated via a single Cargo feature pinning one major
version and a backend major-version bump becomes a Basin major-version
bump. This differs from argmin, which uses versioned backend traits and
requires a new trait for each backend version. We opted to keep the
backend traits versionless and instead version the entire crate in order
to improve maintainability.

\subsection{Compile-Time Correctness}\label{compile-time-correctness}

Generic stopping conditions (iteration limits, tolerance families,
evaluation budgets, and wall-clock limits) are configured uniformly on
the \texttt{Executor} rather than reimplemented per solver, and each
criterion binds on the minimum state shape it needs. This is what makes
an ill-typed pairing (a gradient tolerance on a gradient-free method) a
compile error. The cost of this is more complex generic signatures, but
the benefit is that Basin users can be confident that their stopping
criteria are compatible with their solver and problem.

\subsection{Constraints}\label{constraints}

Constraints describe the \emph{problem}, so in Basin they exist as
problem-side traits rather than in executor configuration or on the
state. Solvers declare the constraints they consume through those
traits, which means that an unconstrained problem handed to a solver
that requires constraints does not compile. For the common case of
reusing an unconstrained solver, opt-in adapters (a log-barrier method
and an augmented-Lagrangian method) wrap the \emph{problem}; each
adapter consumes the constraint trait and exposes only
\texttt{CostFunction} and \texttt{Gradient}, which is precisely what
routes a constrained problem onto an unconstrained solver.

\subsection{Compatiblitiy}\label{compatiblitiy}

Basin is WebAssembly-compatible by default. Parallelism and BLAS/LAPACK
integration are opt-in features, and default paths use a
WebAssembly-safe time shim and a seedable, WebAssembly-safe random
number generator.

The minimum supported Rust version is kept deliberately low in order to
comply with the toolchain requirements of the R package network
\href{https://cran.r-project.org}{CRAN} in order to facilitate Basin's
use in R packages such as
\href{https://cran.r-project.org/package=eulerr}{eulerr}.

\subsection{Scalar Generics}\label{scalar-generics}

The interface is generic over the scalar type, with \texttt{f64} as the
default so existing call sites resolve unchanged, while \texttt{f32}
works across states, solvers, termination criteria, and the math layer.

\section{Research Impact Statement}\label{research-impact-statement}

Basin is used as the optimizer for the Rust library
Eunoia~(\citeproc{ref-larsson2026a}{Larsson \& Gustafsson, 2026}), which
in turn is used in the R package
eulerr~(\citeproc{ref-larsson2018}{Larsson \& Gustafsson,
2018}).\footnote{These packages are also made by the author.} It is also
used in the R package balancing~(\citeproc{ref-barrett2026}{Barrett,
2026/2026}), which calculates optimization-based balancing weights for
causal inference.

Benchmarks against competitors are available at
\href{https://basin.rs/benchmarks}{basin.rs}, showing that Basin
generally outperforms argmin and nlopt and is on par with gomez.

At the time of writing, the crate has been downloaded roughly 30,000
times on \url{https://crates.io/crates/basin} over the last three months
and has been featured in \emph{This Week in
Rust}~(\citeproc{ref-arlynx2026}{Arlynx, 2026}).

\section{AI Usage Disclosure}\label{ai-usage-disclosure}

Generative AI tools were used substantially during the development of
Basin: Claude Code, running Claude Opus 4.8, Claude Opus 5, and Fable 5,
was used for code generation and refactoring, writing unit tests,
writing documentation, and reviewing this manuscript. The author made
all core design decisions---the architecture, the design tenets, and the
API---and reviewed, edited, and validated all AI-assisted contributions.
In order to further verify correctness, the Powell-family solvers were
developed against PRIMA~(\citeproc{ref-zhang2023}{Zhang, 2020/2023}) and
cross-validated against it numerically, and the L-BFGS-B implementation
was checked for numerical agreement with the original Fortran
code~(\citeproc{ref-zhu1997}{Zhu et al., 1997}). The remaining solvers
are covered by a test suite problems.

\section{Acknowledgements}\label{acknowledgements}

As we have mentioned, Basin owes a substantial intellectual debt to
\texttt{argmin}~(\citeproc{ref-kroboth2025}{Kroboth, 2018/2025}). The
Powell-family derivative-free solvers are derived from
PRIMA~(\citeproc{ref-zhang2023}{Zhang, 2020/2023}), Zaikun Zhang's
modern-Fortran reference implementation of M. J. D. Powell's methods,
used as the authoritative source for the exact formulas and as a
cross-validation oracle. The bound-constrained L-BFGS-B solver is a port
of the L-BFGS-B version 3.0 Fortran code by Ciyou Zhu, Richard H. Byrd,
Peihuang Lu, and Jorge Nocedal, with the improvements by José Luis
Morales and Jorge Nocedal~(\citeproc{ref-morales2011}{Morales \&
Nocedal, 2011}). Both are distributed under the BSD 3-Clause License,
and their notices are retained in the Basin source tree.

\section*{References}\label{references}
\addcontentsline{toc}{section}{References}

\phantomsection\label{refs}

\end{document}